\documentclass[sigplan,10pt,screen,nonacm]{acmart}

\renewcommand\footnotetextcopyrightpermission[1]{}
\AtBeginDocument{%
  }

\usepackage{booktabs}
\usepackage{multirow}
\usepackage{array}
\usepackage{xcolor}
\usepackage{xspace}
\usepackage{listings}
\usepackage{enumitem}
\usepackage{balance}

\newcommand{\sysname}{Rollplex\xspace}

\newcommand{\PHB}[1]{\noindent\textbf{#1}\hspace{.5em}} 
\newcommand{\mainref}[1]{\ref{#1}}
\title{\sysname: Cross-Phase GPU Spatial Sharing for Vision Language Model Post-Training}

\author{Hanfeng Lu$^*$}
\affiliation{\institution{HKUST}\country{Hong Kong SAR}}
\author{Tianyu Feng$^*$}
\affiliation{\institution{HKUST}\country{Hong Kong SAR}}
\author{Suyi Li}
\affiliation{\institution{HKUST}\country{Hong Kong SAR}}
\author{Yuheng Zhao}
\affiliation{\institution{HKUST}\country{Hong Kong SAR}}
\author{Wei Gao}
\affiliation{\institution{HKUST}\country{Hong Kong SAR}}
\author{Shaopan Xiong}
\affiliation{\institution{Alibaba Inc}\country{China}}
\author{Ju Huang}
\affiliation{\institution{Alibaba Inc}\country{China}}
\author{Siran Yang}
\affiliation{\institution{Alibaba Inc}\country{China}}
\author{Jiamang Wang}
\affiliation{\institution{Alibaba Inc}\country{China}}
\author{Lin Qu}
\affiliation{\institution{Alibaba Inc}\country{China}}
\author{Wei Wang}
\affiliation{\institution{HKUST}\country{Hong Kong SAR}}

\begin{document}

\begin{abstract}
Vision-language models (VLMs) enable embodied agents to reason and act from visual observations and language instructions. 
Reinforcement learning (RL) post-training enhances these capabilities using task feedback, but current on-policy RL runtimes execute rollout, reference scoring, and actor training in strict serial phases. 
While effective for text-only RL, this phase-granular execution is wasteful for VLMs, where processing dense video inputs and prompt prefixes occupies a large fraction of each phase. 
Because prefix processing is independent of the generated response, it can be run alongside rollout decoding, which leaves GPU compute capacity underutilized, without breaking synchronous on-policy semantics.

We present \sysname, a runtime that decomposes the reference and training phase and moves the prefix computation into the rollout
decode window.  Realizing this schedule requires more than concurrent
kernel launches: naive colocation of Qwen2.5-VL-32\,B requires roughly
165\,GiB per GPU, while rollout and training prefer different
tensor-parallel (TP) degrees and weight layouts.  \sysname addresses
these constraints with two mechanisms.  Phase-aware memory management
controls HBM residency according to producer--consumer lifetimes.
Parallelism-aware weight sharing uses the same physical
storage for layout-compatible tensors across distinct TP degrees and
reconstructs only incompatible tensors, avoiding a complete second
actor copy.  On 32 H800 GPUs, \sysname achieves
$1.23\times$--$1.30\times$ speedup over serial colocation and
$1.57\times$--$2.24\times$ over disaggregation under the same GPU budget,
while preserving the synchronous RL update.
\end{abstract}

\maketitle
\renewcommand{\thefootnote}{\fnsymbol{footnote}}
\footnotetext[0]{$^*$ These authors contributed equally to this work. The order is determined randomly.}
\pagestyle{plain}

\section{Introduction}
\label{sec:intro}

\begin{figure}[t]
    \centering
    \includegraphics[width=0.99\linewidth]{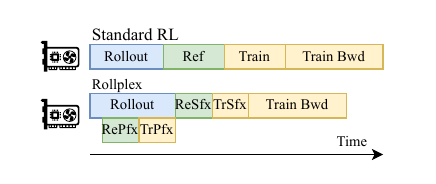}
    \caption{Serial execution and \sysname's cross-phase spatial
    schedule.  RePfx/ReSfx denote the reference-model prefix/suffix;
    TrPfx/TrSfx denote the training-actor prefix/suffix.}
    \label{fig:workflow-example}
\end{figure}

Vision-language models (VLMs) connect perception with language-level
reasoning, making them central to embodied artificial intelligence
(AI) and robotics.  For example, PaLM-E
grounds language in sensor observations, while RT-2 and OpenVLA extend
VLMs into policies that map observations and instructions to robot
actions~\cite{palme,rt2,openvla}.  Efficient post-training on long
visual trajectories is therefore increasingly important.

Current on-policy reinforcement learning (RL) systems for large
language models (LLMs) treat rollout, reference scoring, and actor
training as atomic scheduling phases.  Each iteration generates
responses with an actor snapshot $\theta_k$ (\emph{rollout}), scores
the completed sequences with a frozen reference model, and executes
an actor forward, backward, and optimizer step before publishing
$\theta_{k+1}$.  Production systems~\cite{deepspeedchat,openrlhf,verl,
roll,nemoaligner} distribute these phases across GPUs, but serialize
them to preserve the synchronous on-policy dependence.  This
phase-level schedule is natural when response generation dominates
the iteration. However, it becomes unnecessarily coarse
for input-heavy VLM RL.

VLM RL changes the resource and time profile of this pipeline.
A VLM first encodes video frames into visual tokens and then
processes those tokens together with the text prompt through an LLM
backbone~\cite{qwen25vl,llavavideo,videor1}.  Across the four video
workloads in this paper, prompt tokens account for $79\%$--$98\%$ of
each sampled sequence at the median, compared with $11\%$--$19\%$ for
the text-reasoning workloads MATH and GSM8K
(\S\ref{sec:workload-time}).  Prompt-side computation is therefore no
longer a small tail: it rivals the autoregressive decode that dominates
text-only RL.  Although rollout, reference scoring, and actor training
perform different computations, each contains substantial prompt-side
work.  The reference and actor-training prefixes are required by the
RL objective, but neither depends on the response that rollout is still
generating.  Conventional phase ordering nevertheless delays both
prefixes until rollout completes.

These prefixes are ready before rollout finishes, but phase-level
scheduling prevents them from running.  This mismatch exposes an
opportunity to accelerate the RL iteration: overlap the
rollout-independent prefixes with rollout decode, as
Figure~\ref{fig:workflow-example} illustrates (\S\ref{sec:obs}).  The
reference and training-actor prefixes execute while rollout generates
the response, and their response-dependent suffixes continue after
generation completes.  Because this overlap changes only the execution
order, it preserves synchronous on-policy semantics without speculating
on the response or using a stale policy.

However, existing RL runtimes do not expose this scheduling
granularity.
Colocated systems execute all phases sequentially on a shared GPU
pool~\cite{verl,roll}.  Disaggregated systems place phases on separate
GPU pools but preserve the same sequential execution
order~\cite{openrlhf,rollmux,streamrl}.  Asynchronous and tail-oriented
systems increase rollout throughput by relaxing ordering or selecting
which responses enter an update~\cite{areal,relax,rollpacker,april}.
These techniques improve placement, synchronization, or the decode
long tail, but none executes response-independent work from a later
phase in the spatial slack of the same synchronous rollout.

Naively launching these phases concurrently creates two feasibility
challenges and one performance constraint.  First, overlap extends
tensor lifetimes:
rollout KV state, prefix KV caches, training activations, parameters,
gradients, and optimizer state together require roughly 165\,GiB per
GPU for Qwen2.5-VL-32\,B, more than twice an H800's 80\,GB capacity
(\S\ref{sec:challenge-intermediate}).  Dropping preserved prefix state
saves memory but recomputes the work that overlap tries to hide.
Second, rollout and training prefer different tensor-parallel (TP)
degrees (\S\ref{sec:challenge-tp}).  Training uses a wider group to
distribute model and optimizer state~\cite{megatron,zero}, whereas
rollout prefers a narrower group to reduce per-token
collectives~\cite{vllm,popeefficientinfer}.  A common degree either
exceeds training memory or slows decode; separate actor instances
duplicate weights and require layout conversion after every update.
Third, prefix kernels can contend with decode for SM capacity, cache,
and HBM bandwidth.  Because later computation waits for the generated
response, any decode slowdown extends the critical path and can erase
the benefit of overlap.  The scheduler must therefore control
interference rather than maximize aggregate utilization
(\S\ref{sec:design-overlap}).

We present \sysname, a runtime that makes phase-level RL scheduling
finer grained by decomposing later phases at the first response
position (\S\ref{sec:design}).  It executes their prefixes during
rollout decode, treats decode as the critical path, and
preserves their boundary state for the response-dependent suffixes.
Two mechanisms enable this schedule:

\PHB{Phase-aware memory management.}
\sysname classifies intermediate state by producer, consumers, and
last use.  It retains latency-critical boundary KV state,
offloads or recomputes bulky training state, releases phase-local
rollout and scoring buffers at their last use, and streams the 32-bit
floating-point (FP32) optimizer update in chunks.  A CUDA-VMM-backed
allocator preserves
virtual addresses while changing physical residency, allowing the
working set to move through HBM without rebuilding engine-level tensor
objects.

\PHB{Parallelism-aware weight sharing.}
\sysname backs training and rollout with the same physical actor
storage where their layouts can be represented as aliases.  It
classifies tensors into identical-sharding, transpose-compatible, and
sharing-incompatible cases.  The first two cases share storage
directly or through a transposed logical view; only incompatible
tensors are copied and permuted after an update.  Thus the engines may
use distinct TP degrees without materializing a complete second actor
copy.

We implement \sysname in ROLL~\cite{roll}, with Megatron-Core for
training and vLLM for rollout, and evaluate Qwen2.5-VL-32\,B on
32 H800 GPUs.  Across four video-reasoning workloads, \sysname achieves
$1.23\times$--$1.30\times$ speedup over serial colocation and
$1.57\times$--$2.24\times$ over disaggregation under the same GPU budget
(\S\ref{sec:eval-e2e}).

This paper makes the following contributions:
\begin{enumerate}[leftmargin=*,itemsep=1pt,topsep=2pt]
    \item We characterize VLM RL as an input-heavy regime in
    which phase-level serialization hides useful parallelism, and
    formulate the prefix dependence and decode-window bound of a
    semantics-preserving cross-phase schedule.
    \item We design phase-aware memory management that retains critical
    boundaries in HBM and moves state by producer--consumer lifetime.
    We also develop parallelism-aware weight sharing that classifies
    TP layouts, shares compatible storage, and reconstructs only
    incompatible tensors.
    \item We implement \sysname in ROLL and demonstrate end-to-end
    gains over serial colocation and disaggregation without relaxing
    the synchronous on-policy update.
\end{enumerate}

\section{Motivation and Challenges}
\label{sec:workload}
\label{sec:existing}

\subsection{Synchronous VLM RL}
\label{sec:workload-iter}

One synchronous on-policy iteration uses an actor snapshot
$\theta_k$ in four ordered steps (Figure~\ref{fig:workflow-example}).
The rollout engine first samples responses.  A frozen reference model
then scores the completed sequences to provide the KL regularizer.
The training actor computes the RL loss and runs forward and backward
passes.  Finally, the optimizer updates the actor weights for the next
iteration.
DeepSpeed-Chat, OpenRLHF, verl, ROLL, and NeMo-Aligner differ in
placement and communication, but follow this dependence
structure~\cite{deepspeedchat,openrlhf,verl,nemoaligner,roll}.  VLM RL
adds a vision transformer (ViT) before the language model.  The ViT
converts video frames into visual tokens, which the language model
processes together with the text prompt.  This visual path changes the
time profile of each RL phase.

\subsection{An Input-Heavy Workload}
\label{sec:workload-time}

\begin{figure}[t]
    \centering
    \includegraphics[width=\linewidth]{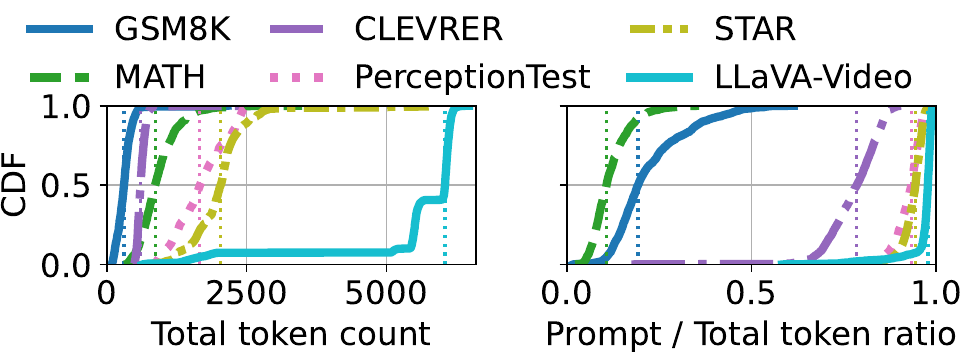}
    \caption{Token statistics for four Video-R1 tasks and the
    text-RL baselines GSM8K and MATH. \textbf{Left}: CDF of total
    tokens per sample. \textbf{Right}: CDF of the prompt fraction.}
    \label{fig:workload-profile}
\end{figure}

Text-only RL is usually decode dominated: reference and reward
inference together account for less than $15\%$ of step
time~\cite{rollpacker,april,rollmux}.  The video-intensive VLM
workloads change this balance for two reasons.  First,
visual prompts are long.  Even after
dynamic-resolution processing, a Qwen-VL encoder at 720p emits roughly
$1{,}536$ visual tokens per 16-frame block~\cite{qwen25vl}.  Second,
the vision encoder and the following long-prompt prefill are dense,
compute-intensive operations absent from text-only RL.

Figure~\ref{fig:workload-profile} quantifies the resulting input-heavy
profile.  Median prompts contain roughly $470$--$5{,}900$ tokens on
the four video tasks, versus 60 and 90 on GSM8K and MATH.  Prompt
tokens constitute $79\%$--$98\%$ of a video sample at the median, but
only $11\%$--$19\%$ of the text baselines.  Because each prompt passes
through rollout prefill, reference scoring, and actor training, its
cost rivals the autoregressive decode that dominates text RL.
Consequently, accelerating decode alone leaves a substantial fraction
of a VLM RL iteration untouched.  This motivates overlapping
prompt-side work with, rather than after, rollout decode.

\subsection{Observation and Opportunity}
\label{sec:obs}
\label{sec:obs-prefix}

We observe that VLM RL draws training data from two sources: visual
inputs and prompts supplied by the dataset, and responses generated
during rollout.  This boundary splits each model invocation at the
first response position into a \emph{prefix} and a \emph{suffix}.
Processing the prefix runs the vision encoder and prompt prefill,
producing boundary KV state and prefix log-probabilities.  Rollout
generates the suffix autoregressively; the reference model scores it,
and the training phase computes gradients for the model update.  The
reference and training-actor prefixes depend only on the input and the
current model weights, not on the response tokens currently being
sampled.  They can therefore run before rollout completes without
changing the response or the synchronous update.  The
response-dependent suffixes must wait for generation, but can continue
from the retained prefix boundary rather than recomputing the prompt.
This independence creates an opportunity to overlap prefix computation
with rollout decoding.

\label{sec:obs-decode}
\label{sec:motivation-slack}

\begin{figure}[t]
\centering
\includegraphics[width=\linewidth]{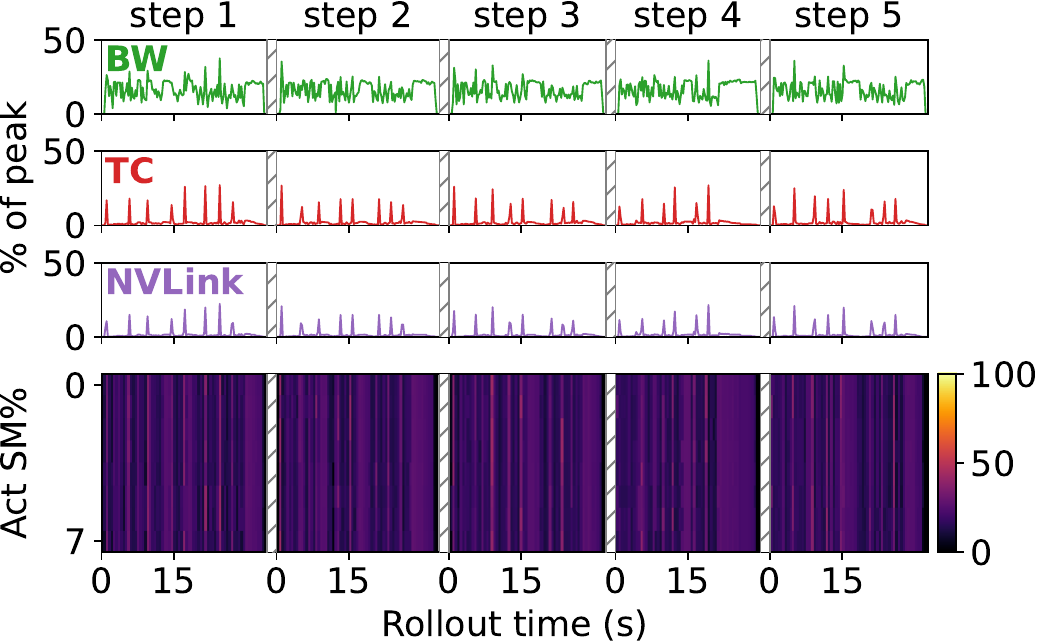}
\caption{Per-GPU utilization during rollout. TC: Tensor Core; BW:
HBM bandwidth.}
\label{fig:sm-utilization}
\end{figure}

Prefix computation is well suited to this overlap because its
utilization profile complements rollout decode.
Token-by-token decoding repeatedly moves weights and KV blocks while
performing little arithmetic per byte, and therefore leaves compute
capacity unused~\cite{vllm,distserve,flashattention}.  In our H800
profile, active-SM utilization remains below $19\%$
(Figure~\ref{fig:sm-utilization}).  Video encoding and prompt prefill,
in contrast, expose dense batched computation.  This complementarity
suggests the cross-phase schedule in
Figure~\ref{fig:workflow-example}: execute the reference and
training-actor prefixes alongside rollout decode, retain their
boundary state, and run only the response-dependent suffixes after
generation.

The schedule preserves strict on-policy ordering because every
overlapped prefix uses the same model weights and input as the serial
schedule.  Its
benefit is capped by the decode window: at most
$\min(T_\mathrm{decode},T_\mathrm{prefix})$ of prefix work can be
removed from the serial critical path.  This bound provides a more
meaningful target than raw GPU utilization: a co-schedule is useful
only to the extent that it hides required prefix work without
expanding the decode window.

\subsection{Challenges}
\label{sec:existing-naive}

Realizing this schedule creates two feasibility challenges:
cross-phase state exceeds HBM, and rollout and training prefer
different TP layouts.

\subsubsection{Cross-Phase Memory Pressure}
\label{sec:challenge-intermediate}

\begin{table}[t]
\centering
\caption{Per-GPU footprint of naive overlap for Qwen2.5-VL-32\,B.}
\label{tab:memory-footprint}
\small
\setlength{\tabcolsep}{5pt}
\begin{tabular}{@{}llr@{}}
\toprule
Role & Component & GiB \\
\midrule
Training actor       & bfloat16 (BF16) weights      & 7.6  \\
(TP=8)               & FP32 gradient buffer         & 15.1 \\
                     & FP32 master weights          & 15.1 \\
                     & FP32 Adam moments            & 30.3 \\
                     & Activations (one microbatch) & 28.3 \\
                     & Preserved prefix KV          & 16.0 \\
\midrule
Reference (TP=8)     & BF16 weights                 & 7.6  \\
\midrule
Rollout (TP=4)       & BF16 weights                 & 15.1 \\
                     & KV cache                     & 29.0 \\
                     & Runtime buffers              & 2--4 \\
\midrule
\multicolumn{2}{@{}l}{Total (vs.\ 80\,GB H800 HBM)} & $\approx$165 \\
\bottomrule
\end{tabular}
\end{table}

While overlap increases GPU utilization, it also introduces
significant memory pressure.  In sequential execution, rollout and
training use HBM at different times, so only the active phase's model
state and intermediate data need to be resident.  Overlapping the
phases instead requires data from multiple phases to coexist in
limited HBM.  Rollout KV must remain live during decode; reference and
actor boundary KV must survive until suffix computation; training
state must remain available for backward; and parameters, gradients,
master weights, and optimizer moments are needed at different points
in the update.  Naively keeping these objects resident requires roughly
165\,GiB per GPU
(Table~\ref{tab:memory-footprint}), more than twice the 80\,GB capacity
of an H800.

The problem is therefore not only allocation size but residency over
time.  A feasible system must retain latency-critical boundary KV,
offload or recompute bulky autograd state, release rollout and scoring
buffers at last use, and avoid materializing the full optimizer state
at once.  It must do so without severing the training-prefix autograd
path that makes the reordered execution equivalent to the serial one.

\subsubsection{Training/Rollout Layout Conflict}
\label{sec:challenge-tp}

Second, rollout and training run in separate driver contexts and
ordinarily allocate independent actor copies, adding roughly
15\,GiB per GPU.  Sharing one physical copy is complicated because the
engines prefer different tensor-parallel (TP) degrees.  Training uses
a wider group, such as TP=8, to distribute parameters, activations,
gradients, and optimizer
state~\cite{megatron,zero}.  Rollout prefers a narrower group, such as
TP=4, because each transformer block places a collective on the
per-token decode path~\cite{vllm,popeefficientinfer}.  Matched TP=4
does not fit the training state, whereas matched TP=8 slows rollout by
up to $1.31\times$ (\S\ref{sec:eval-ipc}).  An ideal system should
therefore maintain one physical copy of the actor weights, share it
across phases, and still allow rollout and training to use their
preferred TP degrees.

Doing so introduces a further layout problem.  Different TP degrees
induce different shard boundaries, and Megatron and vLLM may order the
same logical parameter differently in memory.  A shared buffer is
useful only if both engines can express their shards as valid views of
the same physical bytes; otherwise the runtime must copy or permute
the incompatible tensors.

Together, these constraints make naive spatial overlap impractical:
it can exceed HBM capacity or force an inefficient common TP degree.
\sysname addresses both barriers without changing the synchronous
update.

\section{\sysname Design}
\label{sec:design}

\sysname realizes the cross-phase schedule of \S\ref{sec:obs} with two
mechanisms.  \emph{Phase-aware memory management}
(\S\ref{sec:data-engine}) moves the iteration's changing working set
through HBM without breaking the training graph.
\emph{Parallelism-aware weight sharing}
(\S\ref{sec:weight-engine}) lets training and rollout use different TP
degrees while sharing compatible actor storage.

\sysname is implemented in ROLL~\cite{roll}, with Megatron-Core
0.13.0 for training and vLLM 0.17.0~\cite{vllm} for rollout.  CUDA MPS
provides cross-process execution concurrency, while CUDA VMM and IPC
provide physical residency control and shared mappings.
\label{sec:design-impl}

\subsection{Cross-Phase Execution}
\label{sec:design-overview}
\label{sec:design-overlap}

\begin{figure*}[t]
\centering
\includegraphics[width=\linewidth]{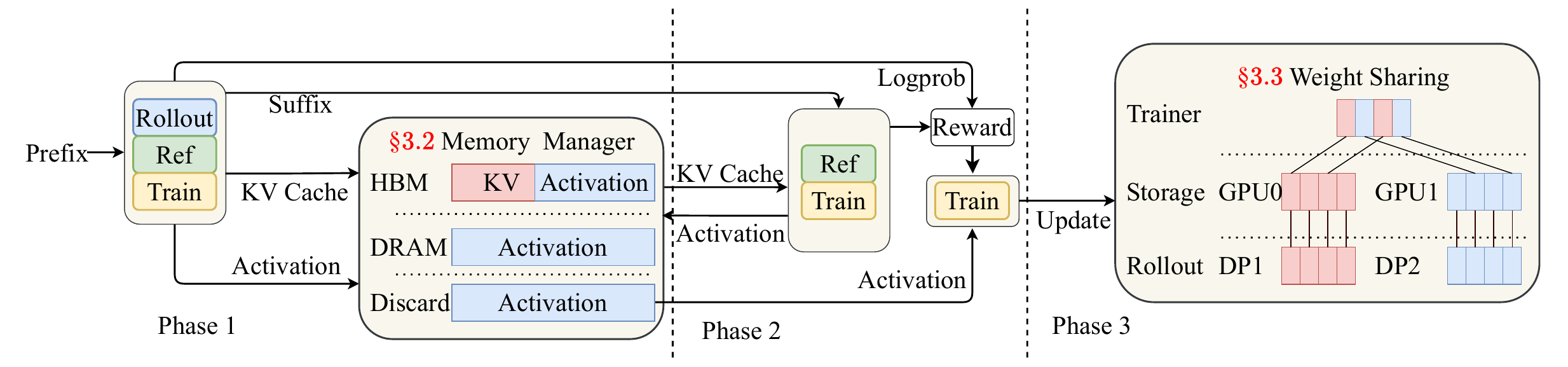}
\caption{\sysname architecture and one cross-phase iteration.  Prefix
work from reference scoring and actor training executes alongside
rollout decode; response-dependent suffixes and backward follow
generation.}
\label{fig:system-overview}
\end{figure*}

\sysname colocates rollout and training as separate processes on one
GPU pool (Figure~\ref{fig:system-overview}).  An orchestrator enforces
phase barriers and treats rollout decode as the critical path.  It
leaves MPS limits unset; the resulting decode interference $\Delta_D$
is evaluated in \S\ref{sec:eval-sm-sharing}.  Prefix kernels run
concurrently on separate process streams; if they outlast decode,
their remainder stays on the critical path.
Every overlapped prefix uses the same input and actor snapshot
$\theta_k$ as the serial iteration, so the schedule changes only
\emph{when} independent work executes.

The orchestrator makes this dependence explicit with three barriers.
During Phase~1, every engine treats $\theta_k$ as read-only.  The
Phase-2 suffixes start only after rollout has fixed the response tokens
and their corresponding prefix boundaries are ready.  The Phase-3
update starts only after reference scoring, actor forward, and backward
have stopped reading $\theta_k$.  These barriers also delimit when
physical pages may be released or rebound: an engine never observes a
mapping change while one of its kernels can still dereference the old
view.

\PHB{Phase 1: rollout and prefix materialization.}
While rollout decodes responses under $\theta_k$, the reference model
and training actor encode the video and prefill the prompt.  Each
produces prefix log-probabilities and boundary KV state for its later
suffix.  Reference outputs are detached.  The training boundary
remains attached to autograd, while bulky saved prefix state is
offloaded or marked for recomputation.  During this window, the actor
snapshot, rollout KV, and boundary state are hot in HBM; optimizer
state is not.

\PHB{Phase 2: suffix scoring and training.}
After generation, the reference and actor suffixes consume the
responses and their preserved boundaries.  Prefix and suffix
log-probabilities form the full-sequence values used by the RL loss,
after which the actor executes backward.  Rollout KV and inference
workspace are released before the training working set reaches its
peak.

\PHB{Phase 3: update and publication.}
The optimizer streams FP32 state through HBM and regenerates the BF16
actor snapshot chunk by chunk.  Training ranks synchronize the
updated pieces and populate any rollout-visible regions required by
the sharing layout.  Because compatible rollout tensors view the same
IPC-backed storage, the completed snapshot is immediately visible for
the next iteration; only incompatible tensors require reconstruction.

Let $T_D$ be standalone decode time, $T_P$ the makespan of the
concurrently running reference and training prefixes, $\Delta_D$ the
decode slowdown caused by co-running them, and $T_\mathrm{post}$ the
response-dependent suffix, backward, and update time.  Ignoring
rollout prefill common to both schedules,
\begin{equation}
\begin{aligned}
T_{\mathrm{Rollplex}} &=
\max(T_D+\Delta_D,\;T_P) + T_\mathrm{post},\\
S &= T_D+T_P-\max(T_D+\Delta_D,T_P)
   =\min(T_P-\Delta_D,T_D).
\end{aligned}
\label{eq:overlap-model}
\end{equation}
Here $S$ is the critical-path saving over the serial interval
$T_D+T_P$.  In the prefix-bounded regime, $S=T_P-\Delta_D$, so
interference reduces the saving linearly; in the decode-bounded regime,
$S=T_D$, and interference is hidden until the boundary.  This
decode-window cap explains why faster decode (\S\ref{sec:eval-ipc}) and
rollout tail-cutting (\S\ref{sec:eval-async}) leave more prefix work on
the critical path.

\subsection{Phase-Aware Memory Management}
\label{sec:data-engine}
\label{sec:design-mem}

\sysname enforces phase-scoped memory residency with one invariant: an
object occupies HBM only between its first latency-critical use and
its last use.  The relevant
objects fall into four classes.  \emph{Permanent state}, such as FP32
master weights and Adam moments, may move between host and GPU but is
never discarded.  \emph{Regenerable snapshots}, including BF16 actor
weights, can be replaced after their last consumer finishes.
\emph{Boundary state}, principally prefix KV, remains resident until
the corresponding suffix consumes it.  \emph{Phase-local state}, such
as rollout KV, inference workspace, scoring temporaries, and
checkpointed activations, is released at last use or recomputed.
\label{sec:design-mem-policy}

\PHB{Stable-address staging.}
\label{sec:design-mem-vmm}
\label{sec:design-mem-pool}
\sysname combines a CUDA-VMM allocator with a pinned host memory pool.  VMM
separates virtual tensor views from their physical backing: at a safe
phase barrier, \sysname can release or remap HBM pages without
rebuilding the Megatron and vLLM tensor objects that refer to those
address ranges~\cite{cudavmm,vattention,gmlake}.  VMM allocations are
also exported through CUDA IPC so both processes can map compatible
actor storage.  The pinned host memory pool holds inactive permanent
state and offloaded saved tensors; allocating these objects from one
pool avoids fragmentation from many large pinned allocations.

The allocator reserves virtual ranges once and maps physical pages
only while an object is resident.  A release operation unmaps and
frees the backing pages while retaining the reservation; remapping
later attaches fresh pages at the same virtual address.  The tensor's
shape, strides, and storage offset
therefore remain valid across an eviction.  For shared actor ranges,
the owner exports VMM allocation handles and the peer process imports
and maps the same physical pages into its own address space.  Imported
mappings persist across iterations unless the underlying allocation is
resized or replaced.  \sysname preallocates the pinned host memory
pool during initialization.

Using VMM and the preallocated host pool, \sysname manages residency
across the three execution phases.  Rollout KV and inference buffers
are released after generation.  Boundary KV
survives only until its Phase-2 suffix.  Frozen reference weights are
paged into HBM for their Phase-1 prefix and Phase-2 scoring windows
and may be evicted between those windows.  Training-prefix saved state
is offloaded as it is produced or discarded for checkpoint
recomputation.  At the optimizer barrier, all consumers of the old
BF16 snapshot have quiesced; \sysname releases its physical pages and
reuses that capacity as update workspace.

Consequently, the Phase-1 peak contains the actor snapshot, rollout KV,
the two prefix boundaries, the reference weights only while their
prefix runs, and small offload metadata.  Suffix activations and
gradients appear only in Phase~2, after rollout state is released.
Full optimizer state never joins either peak; Phase~3 admits only a
sliding window of it.

\PHB{Chunked update.}
\label{sec:design-mem-update}
Loading FP32 master weights and Adam moments for the complete model
would exceed HBM even after releasing the old BF16 snapshot.
\sysname therefore partitions parameters into byte-balanced chunks.
For each chunk it admits the finalized gradient and corresponding
FP32 state, applies Adam, casts the result to BF16, and writes the new
snapshot into the rollout-visible destination selected by
\S\ref{sec:weight-engine}.  Updated FP32 state is returned to the
host, and the workspace is reused by the next chunk.  Adjacent
transfers and computation are pipelined when memory permits.  Global
loss-scale, norm, and clipping operations complete before any affected
chunk is committed, so this changes residency rather than optimizer
semantics.

The implementation uses separate load, update, and offload streams over
adjacent chunks.  While chunk $j$ executes Adam, chunk $j+1$ may load
its FP32 state and chunk $j-1$ may return its updated state to host.
Chunk size is selected so this small pipeline, the partially regenerated
BF16 snapshot, and the current gradient slice fit simultaneously.
Peak update memory is therefore independent of the full FP32 optimizer
footprint.

\PHB{Training-prefix correctness.}
\label{sec:design-mem-contract}
Offloading saved tensors does not detach the prefix computation.
The actor prefix produces attached boundary KV; the suffix loss sends
gradients through that boundary into the LLM prefix, projector, and
vision encoder.  Saved prefix tensors are reloaded or recomputed when
backward needs them.  Consequently, \sysname preserves the serial
autograd path while keeping only latency-critical boundary state in
HBM across the decode window.  The supplementary lifecycle table
lists the complete per-object contract.  The reference prefix has no
autograd obligation and is detached; its retained boundary is consumed
only by reference scoring.

\subsection{Parallelism-Aware Weight Sharing}
\label{sec:weight-engine}
\label{sec:design-ipc}

Training and rollout require the same logical actor but prefer
different TP degrees and may use different parameter orders.  \sysname
chooses one VMM-backed physical placement for each compatible tensor,
exports it through CUDA IPC, and binds each engine's local tensor to
the appropriate region.  When
$\mathrm{TP}_{\mathrm{train}}>\mathrm{TP}_{\mathrm{rollout}}$,
several training shards jointly cover one rollout shard; after an
update, ranks exchange the pieces needed to populate each
rollout-visible shared region.

Let $q=\mathrm{TP}_{\mathrm{train}}/\mathrm{TP}_{\mathrm{rollout}}$.
The current sharing scheme applies when $q\in\mathbb{Z}_{>0}$.
\sysname groups every $q$ colocated training ranks
whose shards form one rollout shard.  The rollout-sized IPC buffer is
the actor snapshot itself rather than a separate post-update copy.
Each training rank binds its parameter to the region it owns and writes
its updated BF16 values there.  A group-local exchange gathers the
other $q-1$ pieces and scatters them to their predetermined offsets so
that every rollout rank sees a complete shard.  Layout classification
determines whether this population is a direct placement, a
metadata-compatible transpose, or an explicit reconstruction.

\begin{figure}[t]
\centering
\includegraphics[width=0.99\linewidth]{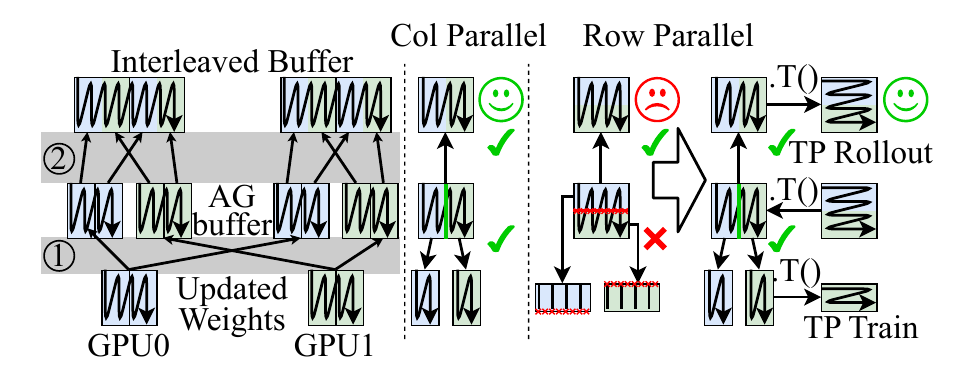}
\caption{Weight sharing for
$\mathrm{TP}_{\mathrm{train}}=2\mathrm{TP}_{\mathrm{rollout}}$.
\textbf{Left}: training ranks populate a rollout-sized shared region.
\textbf{Middle}: same-axis shards alias directly. \textbf{Right}:
row-parallel shards share a transposed storage order.}
\label{fig:weight-sharing}
\end{figure}

\PHB{Sharing criterion.}
For logical element $i$ of a tensor, let $\phi_E(i)$ denote the
physical byte observed by engine $E$ after metadata-only transforms
accepted by its kernels.  A tensor is zero-copy shareable when a
common physical placement makes
\[
    \phi_{\mathrm{train}}(i)=\phi_{\mathrm{rollout}}(i)
    \quad\text{for every }i,
\]
and each rank-local shard satisfies the engine's contiguity
requirements.  VMM operates on page-aligned imported ranges; tensor
views select offsets within those ranges.  \sysname tests the
criterion using the tensor's sharding axis, shard order, storage
strides, and TP ratio, producing three cases.

This classification is performed once when the engines bind their
parameters.  Equality of the two mappings guarantees that a training
write changes exactly the bytes later read by rollout.  When no
permitted metadata transform makes the mappings equal, \sysname marks
the tensor for reconstruction rather than exposing an incorrect alias.

\PHB{Case 1: identical or nested sharding.}
\label{sec:design-ipc-uniform}
If both engines shard the same axis in the same order, every training
shard is a contiguous region of the rollout placement.  This includes
identical TP and nested partitions in which the wider training TP
subdivides a rollout shard, as well as replicated vision parameters.
Both engines bind views of the common allocation; no layout
conversion is required.

\PHB{Case 2: transposed storage.}
Some row-parallel weights fail Case 1 only because the sharded axis is
innermost in row-major storage.  \sysname stores such a weight in the
transposed order, making each shard a contiguous slab, and exposes the
expected logical orientation through tensor metadata.  The physical
bytes remain shared; GEMM kernels use their transpose mode.  Concretely, a logical
$[O,I]$ matrix sharded along $I$ is stored as $[I,O]$ so every
training and rollout slice occupies a contiguous range; neither engine
materializes the logical transpose.

\PHB{Case 3: incompatible permutation.}
\label{sec:design-ipc-recon}
A non-affine permutation cannot satisfy the common-placement
criterion.  For example, vLLM and Megatron can order fused-QKV heads
differently (interleaved versus grouped).  The reordered
fragments are smaller than the 2\,MB VMM page granularity and cannot
be reconciled by page remapping or a single strided view.  \sysname
therefore keeps a separate inference-layout buffer only for these
tensors and copies and permutes them after an update.

\PHB{Binding and refresh.}
\label{sec:design-ipc-refresh}
The streaming update writes Case-1 and Case-2 BF16 values into shared
destinations and reconstructs only Case 3; Case 2 uses the common
transposed order.
Case-3 reconstruction starts after its training chunk is final and
finishes before rollout binds the new snapshot.  The Phase-3 barrier
prevents either engine from mixing chunks from $\theta_k$ and
$\theta_{k+1}$.  Mappings persist across iterations; before remapping
or resizing a buffer, the orchestrator quiesces both engines.  This
avoids a second actor while preserving each engine's preferred TP
degree.


\section{Evaluation}
\label{sec:eval}

We evaluate \sysname by answering five questions:
\textbf{Q1:} How much does prefix--decode overlap reduce end-to-end
step time over colocated and disaggregated deployments?
(\S\ref{sec:eval-e2e})
\textbf{Q2:} Does the new schedule change what is
learned? (\S\ref{sec:eval-reward})
\textbf{Q3:} Are \sysname's two techniques necessary for the
overlap: does phase-aware memory management keep the schedule within
HBM (\S\ref{sec:eval-mem}), and how much performance does
parallelism-aware weight sharing provide (\S\ref{sec:eval-ipc})?
\textbf{Q4:} How should the co-running processes share GPU execution
resources? (\S\ref{sec:eval-sm-sharing})
\textbf{Q5:} Does \sysname remain effective on top of asynchronous
rollout-side optimizations? (\S\ref{sec:eval-async})

\subsection{Experimental Setup}
\label{sec:eval-setup}

\PHB{Models.}
Unless otherwise stated, we train Qwen2.5-VL-32\,B~\cite{qwen25vl}. All
training follows Video-R1's reinforcement learning with verifiable
rewards (RLVR) setup~\cite{videor1}, using Group Relative Policy
Optimization (GRPO)~\cite{grpo} with $N=8$ responses per prompt group. Unless
otherwise stated, the maximum sequence length is 10k tokens, split into
an 8k-token prompt budget and a 2k-token response budget, and the
rollout batch size is 32 prompts.

\PHB{Cluster Setup.}
We evaluate \sysname on a 4-node H800 cluster with 32 GPUs in total,
1.4 TB of host memory per node, and 400 Gbps InfiniBand.
The memory microbenchmarks use a single 8-GPU H800 node.  The
resource-policy study of \S\ref{sec:eval-sm-sharing} uses
8$\times$H20 (78 SMs per GPU).

\begin{sloppypar}
\PHB{Datasets.}
We use CLEVRER~\cite{clevrer},
Perception\allowbreak Test~\cite{perceptiontest},
LLaVA-\allowbreak Video-178K~\cite{llavavideo}, and STAR~\cite{star}. These tasks
span short synthetic physical reasoning clips, real-world video
understanding, long-video question answering, and situated action
reasoning.
\end{sloppypar}

\PHB{Measurement.}
After discarding the first iteration for engine warmup and CUDA-graph
capture, we report average time per step over the remaining iterations.

\subsection{End-to-End Evaluation}
\label{sec:eval-e2e}
\label{sec:eval-overlap}

We compare the end-to-end performance of \sysname with two major
deployment paradigms:
\begin{enumerate}
  \item \textbf{Colocate.} Rollout and training share the same GPU pool,
  but phases execute sequentially, following the common colocated ROLL
  execution style~\cite{roll}.
  \item \textbf{Disagg.} Rollout uses two dedicated 8-GPU nodes with
  \texttt{TP=4}, while training uses two dedicated 8-GPU nodes with
  \texttt{TP=8}. This preserves the same 32-GPU budget and requires
  cross-pool actor-weight synchronization after each update.
\end{enumerate}

We implement every system in ROLL~\cite{roll}. Each enables the same
prefix-sharing optimization~\cite{prefixgrouper,areal_dta,tree_training}.
All systems use the same model, dataset order, reward function,
rollout batch size, and maximum sequence lengths.



\begin{figure}[t]
\centering
\includegraphics[width=\linewidth]{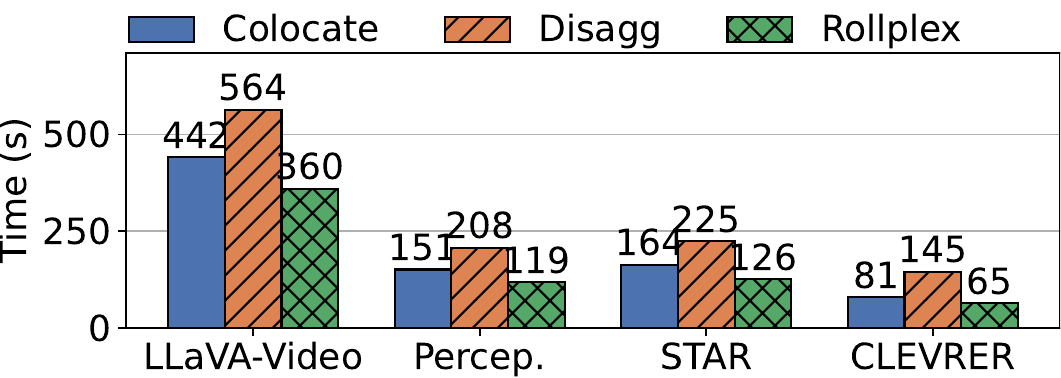}
\caption{Per-step execution time of Colocate, Disagg, and \sysname{}
on Qwen2.5-VL-32\,B over 32$\times$H800.}
\label{fig:e1-e2e}
\end{figure}

As shown in Figure~\ref{fig:e1-e2e}, \sysname delivers the lowest step time on every dataset. It reduces
step time by $1.23\times$--$1.30\times$ over Colocate and by
$1.57\times$--$2.24\times$ over Disagg. The gain over Colocate comes from moving
response-agnostic prefix computation into the rollout-decode
window, while \sysname's memory and weight-sharing mechanisms keep the
overlapped working set within HBM.

The gain is consistent across datasets,
and its magnitude reflects the decode-window cap of \S\ref{sec:obs}:
overlap can hide at most the shorter of decode time and total prefix
time, while the response-dependent suffixes, the backward pass, and
the model update remain on the critical path. Hiding the prefix work
alone therefore yields around $1.2\times$--$1.3\times$ end-to-end.

Disagg trails Colocate on every dataset despite using the same 32-GPU budget: with dedicated
pools, the rollout pool idles during scoring and training while the
training pool idles during rollout, and per-iteration cross-pool
weight synchronization adds further overhead. The cost is
proportionally largest on CLEVRER ($1.80\times$), whose short
iterations amortize these fixed costs least.

\subsection{End-to-End Reward Comparison}
\label{sec:eval-reward}

The speedup of \S\ref{sec:eval-e2e} is only meaningful if training
quality is unchanged.
The systems techniques in \sysname change when computation runs, not
the GRPO update being computed: the overlapped prefixes are computed
under the same on-policy snapshot $\theta_k$, and the preserved prefix
KV is exactly the state the serial schedule would recompute. To
confirm this empirically, we train with \sysname and the colocated
baseline under the same model, data order, reward function, and
response budget, and compare the per-step verifiable reward of the
two systems on all four datasets.

\begin{figure}[t]
\centering
\includegraphics[width=\linewidth]{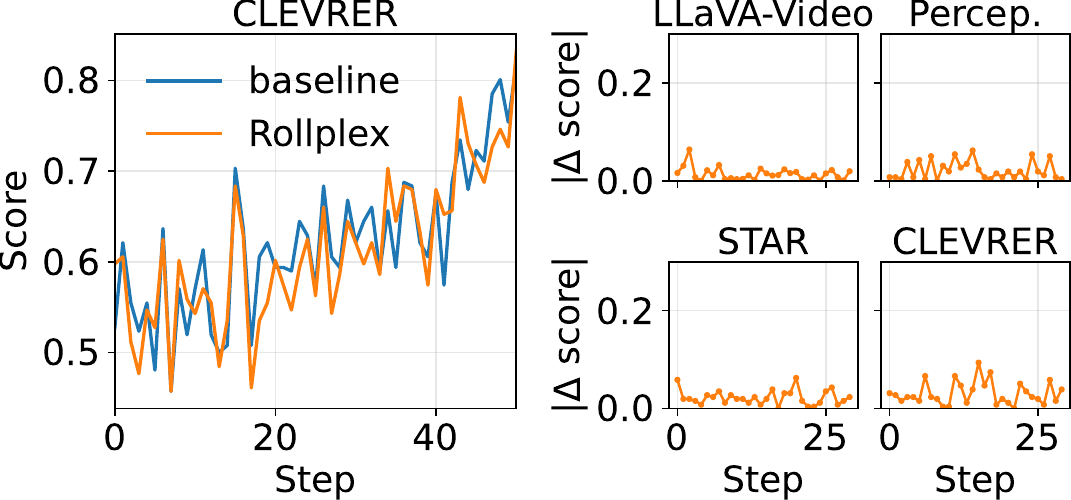}
\caption{Training-quality comparison. \textbf{Left}: Verifiable reward per step
for \sysname and the colocated baseline on CLEVRER. \textbf{Right}: Absolute
per-step reward difference between the two systems on all four
datasets.}
\label{fig:e2-reward}
\end{figure}

Exact trajectories need not match because GRPO training is stochastic,
and different GPU kernel choices can introduce small floating-point
differences.  Nevertheless, Figure~\ref{fig:e2-reward} left shows that
the two CLEVRER reward curves closely track each other throughout
training and reach similar final rewards.
Figure~\ref{fig:e2-reward} right plots the absolute per-step reward
difference between the two systems on all four datasets. It stays
near zero across training, with no growth or drift.  This negligible
reward gap is within the normal step-to-step variation of GRPO
sampling. \sysname therefore improves time-to-train without affecting
training quality.

\subsection{Memory Feasibility}
\label{sec:eval-mem}

To fit in HBM, the overlapped schedule uses phase-aware memory
management (\S\ref{sec:data-engine}). On the 8$\times$H800
microbenchmark with the 32\,B model, we profile per-GPU memory for
full \sysname and four ablations: \emph{no state offload} keeps
optimizer state and activations in HBM; \emph{vLLM resident} retains
rollout KV and inference buffers; \emph{unchunked optimizer} loads all
FP32 master weights and Adam state together; and \emph{no weight
sharing} duplicates the actor weights.

\begin{figure}[t]
\centering
\includegraphics[width=\linewidth]{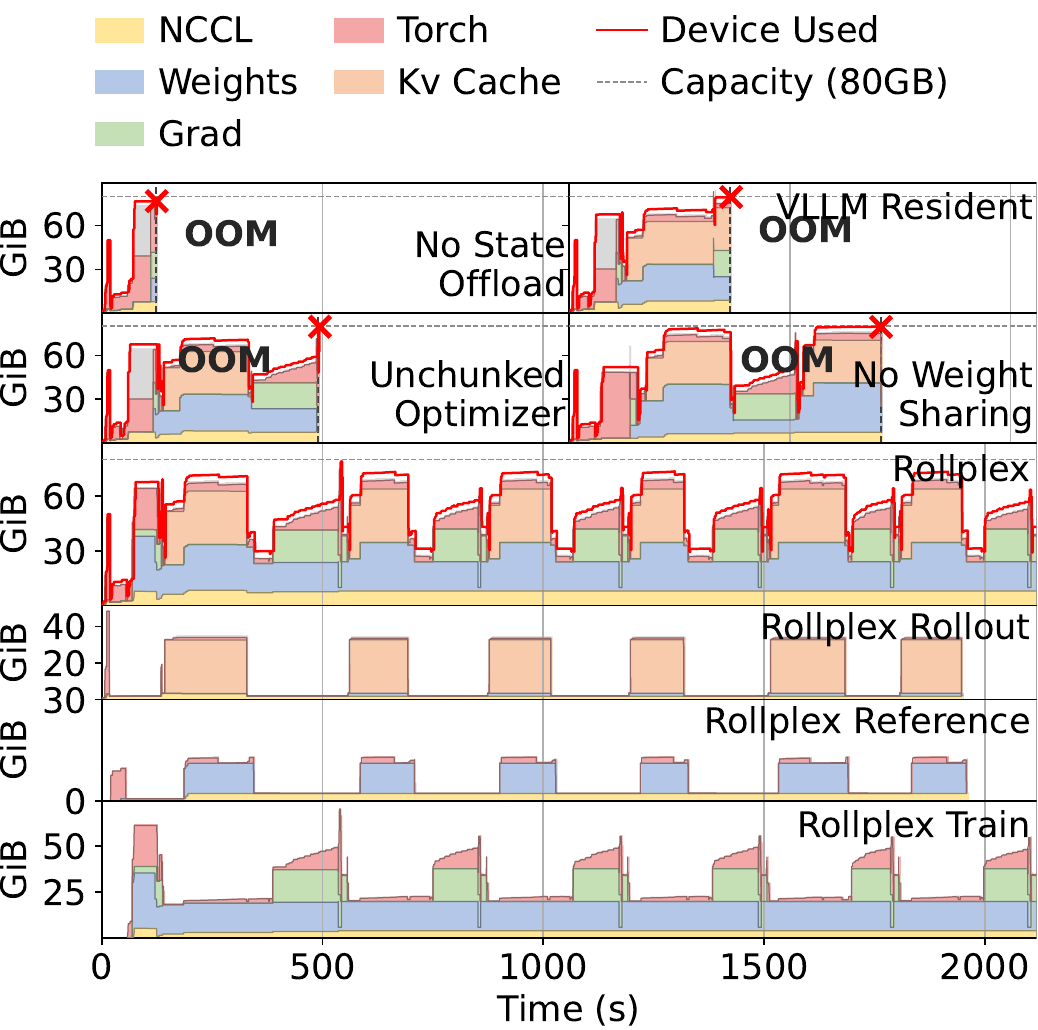}
\caption{Per-GPU memory over time. Top: four ablations; middle: full
\sysname; bottom: per-role breakdown. Dashed line: 80\,GB capacity.}
\label{fig:e3a-mem}
\end{figure}

Figure~\ref{fig:e3a-mem} shows the result: all four ablations run out
of memory, while full \sysname runs continuously with its peak below
the 80\,GB capacity. The failures occur where the lifetime analysis of
\S\ref{sec:challenge-intermediate} predicts. Without state offload,
the working set exceeds capacity already in the first iteration. The
remaining three ablations all die at an optimizer step, the
peak-memory point of the iteration: the unchunked optimizer loads the
full FP32 master weights and Adam state at once and fails immediately.
Keeping vLLM state resident or duplicating the actor weights survives
longer, but the extra resident state eventually collides with the
optimizer's transient peak.
\sysname's chunked optimizer caps exactly this peak, which, combined
with releasing the BF16 weight snapshot at the barrier, is what keeps the
full system under capacity.

The per-role panels illustrate phase-scoped memory residency.
Rollout KV occupies HBM only during the decode windows and is released
after generation. Reference weights are loaded around reference
compute and offloaded as soon as it completes. On the training
process, the BF16 actor snapshot is released at the optimizer
barrier, which appears as the dip in the weights band, so that the streamed
optimizer chunks can reuse that memory before the next snapshot is
written back. Finally, the gradual within-iteration slope in the
stacked areas comes from the PyTorch caching allocator, which retains
freed blocks rather than returning them to the device. This is reused
cache, not live tensor state, and does not reflect actual memory
demand.

\PHB{Weight-layout coverage.}
For Qwen2.5-VL-32\,B, Cases 1/2/3 cover 19.7/10.7/2.4\,B
parameters (60.0/32.8/7.2\%).  Cases 1 and 2 therefore share physical
storage for 92.8\% of parameters.  The 1.1\,GiB Case-3 buffer per GPU
at rollout \texttt{TP=4} is 85.5\% smaller than duplicating the smaller
\texttt{TP=8} actor shard (7.6\,GiB; Table~\ref{tab:memory-footprint}).

\subsection{Tensor-Parallel Degree versus End-to-End Time}
\label{sec:eval-ipc}

We evaluate parallelism-aware weight sharing
(\S\ref{sec:weight-engine}) by comparing rollout at \texttt{TP=8},
which matches training, with its preferred \texttt{TP=4}.  Both use
the 32$\times$H800 setup of \S\ref{sec:eval-e2e}, with training fixed
at \texttt{TP=8}.

\begin{figure}[t]
\centering
\includegraphics[width=\linewidth]{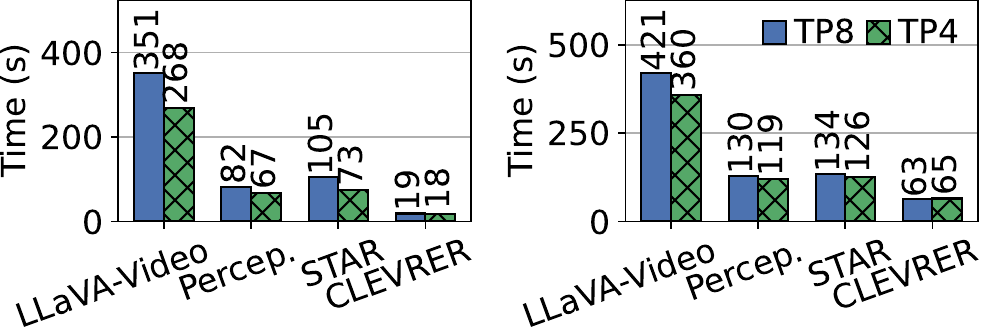}
\caption{Rollout \texttt{TP=8} versus \texttt{TP=4} under \sysname
with training fixed at \texttt{TP=8}. \emph{Left:} rollout time.
\emph{Right:} per-step time.}
\label{fig:e3b-tp}
\end{figure}

Figure~\ref{fig:e3b-tp} (left) shows the cost of the matched
degree. On LLaVA-Video, PerceptionTest, and STAR, \texttt{TP=8} slows
rollout by $1.17\times$--$1.31\times$ relative to \texttt{TP=4},
because the all-reduce on the per-token decode path spans twice as
many ranks. On CLEVRER, whose rollouts are too short for the
communication penalty to accumulate, the two degrees are within a few
percent. The end-to-end picture (right) follows:
rollout \texttt{TP=4} reduces per-step time by
$1.06\times$--$1.17\times$ on the three longer datasets, while on
CLEVRER the two configurations again differ by only a few percent.
The end-to-end gap is also smaller
than the rollout gap because part of the rollout saving is reabsorbed
by the schedule: a faster decode window hides less prefix work, so
some previously overlapped prefix computation returns to the critical
path. This is most visible on STAR, where a $25$-second rollout saving yields
a $7$-second step-time saving.

Letting rollout keep its preferred degree helps wherever rollout
is long enough to matter and costs nothing elsewhere.
Parallelism-aware weight sharing makes this choice free:
rollout runs at \texttt{TP=4} against the \texttt{TP=8} training
layout while both engines read one physical actor weight allocation, with no
duplicate actor copy or separate full-model layout conversion. Without it, a
colocated deployment must either accept the matched-TP slowdown or pay
the duplicate-weight memory cost of \S\ref{sec:challenge-tp}.

\subsection{Resource Sharing Policy}
\label{sec:eval-sm-sharing}

The overlap model of Equation~\ref{eq:overlap-model} depends on keeping
rollout interference $\Delta_D$ small.  We compare default MPS with
explicit MPS limits and GreenCtx partitions in a single-node setup
that isolates GPU-sharing interference from cross-node communication.
The setup uses Qwen2.5-VL-7B on 8$\times$H20, with
Megatron training at \texttt{TP=4} and two vLLM rollout clients at
\texttt{TP=4}.  \emph{ThreadCap} sets per-client active-thread
percentages; \emph{SM-Affinity} converts them to SM-count affinity.
MPS ratios list training followed by each rollout client.
\emph{GreenCtx-Partition} reports actual training/rollout SM counts
and runs without MPS.  All configurations use the same overlapped
schedule.

\begin{table}[t]
\centering
\footnotesize
\setlength{\tabcolsep}{2.4pt}
\renewcommand{\arraystretch}{1.05}
\caption{Resource-sharing policies for Qwen2.5-VL-7B on
8$\times$H20. Each cell reports total step time (top) and rollout
generation time (bottom), in seconds. $^\dagger$Sequential disables
cross-phase overlap and provides the standalone rollout reference.}
\label{tab:smpart-main}
\begin{tabular}{@{}lrrrr@{}}
\toprule
Policy & CLEVRER & LLaVA & Percep. & STAR \\
\midrule
Sequential (no overlap)$^\dagger$ & \shortstack{44.5\\15.0} & \shortstack{216.1\\133.7} & \shortstack{163.4\\79.7} & \shortstack{56.2\\23.9} \\
\cmidrule(lr){1-5}
MPS-Default       & \shortstack{39.2\\16.7} & \shortstack{\textbf{197.8}\\134.9} & \shortstack{115.5\\81.4} & \shortstack{41.1\\23.8} \\
\cmidrule(lr){1-5}
ThreadCap 80\%/20\%   & \shortstack{40.0\\19.0} & \shortstack{199.6\\135.4} & \shortstack{120.8\\84.5} & \shortstack{\textbf{40.7}\\23.6} \\
\cmidrule(lr){1-5}
ThreadCap 70\%/30\%   & \shortstack{126.6\\73.8} & \shortstack{259.5\\177.0} & \shortstack{222.6\\150.1} & \shortstack{91.6\\65.0} \\
\cmidrule(lr){1-5}
SM-Affinity 80\%/20\% & \shortstack{\textbf{38.2}\\17.3} & \shortstack{200.2\\137.3} & \shortstack{115.4\\76.3} & \shortstack{42.2\\24.3} \\
\cmidrule(lr){1-5}
SM-Affinity 70\%/30\% & \shortstack{131.2\\75.5} & \shortstack{259.5\\175.7} & \shortstack{213.5\\140.4} & \shortstack{93.7\\65.5} \\
\cmidrule(lr){1-5}
GreenCtx 62/16 SMs    & \shortstack{42.4\\24.3} & \shortstack{201.0\\138.3} & \shortstack{\textbf{114.6}\\79.7} & \shortstack{43.5\\27.4} \\
\cmidrule(lr){1-5}
GreenCtx 54/24 SMs    & \shortstack{42.3\\22.6} & \shortstack{203.0\\140.6} & \shortstack{115.8\\79.9} & \shortstack{45.8\\27.9} \\
\cmidrule(lr){1-5}
GreenCtx 38/40 SMs    & \shortstack{43.6\\23.1} & \shortstack{199.0\\137.2} & \shortstack{119.0\\82.0} & \shortstack{45.5\\27.6} \\
\bottomrule
\end{tabular}
\end{table}

\PHB{Work-conserving overlap keeps decode interference small.}
Relative to standalone rollout under sequential execution, MPS-Default
adds at most 1.7 seconds of generation time.  This is 0.9\% on LLaVA
and 2.1\% on PerceptionTest; short CLEVRER incurs 11.3\% but only
1.7 seconds in absolute terms, while STAR differs by 0.1 seconds within
measurement noise.  The overlap nevertheless reduces total step time
by $1.09\times$--$1.41\times$.

\PHB{Default MPS is robust without tuning.}
MPS-Default remains within 2.7\% of the best policy on every dataset
without tuning because either process can consume idle capacity.

\PHB{Rigid limits cause a performance cliff.}
Both 80\%/20\% MPS policies remain within 5\% of MPS-Default, but changing
the limits to 70\%/30\% increases total time by
$1.30\times$--$3.44\times$ relative to the corresponding 80\%/20\%
policy.  This nonlinear response is consistent with ZipBatch's
observation that rigid GPU allocations poorly match kernels whose
compute and bandwidth demands vary over time~\cite{zipbatch,sgdrc}.  Fixed
compute limits do not isolate shared HBM bandwidth, so they can
throttle prefix progress while retaining decode contention.
MPS-Default instead lets either process use idle capacity.

\PHB{Static GreenCtx partitions do not consistently improve the joint critical
path.}
GreenCtx executes successfully by itself, but its total time ranges
from 0.8\% better to 11.4\% worse than MPS-Default and it slows
generation most visibly on the short CLEVRER and STAR workloads.
Because its SM groups are static, capacity left idle by one process
cannot be borrowed by another.  We therefore use MPS-Default: it is not the
lowest-time cell for every workload, but it stays close to the best
without a workload-specific partition.

\subsection{Composition with Asynchronous Rollout Optimizations}
\label{sec:eval-async}

\begin{figure}[t]
\centering
\includegraphics[width=\linewidth]{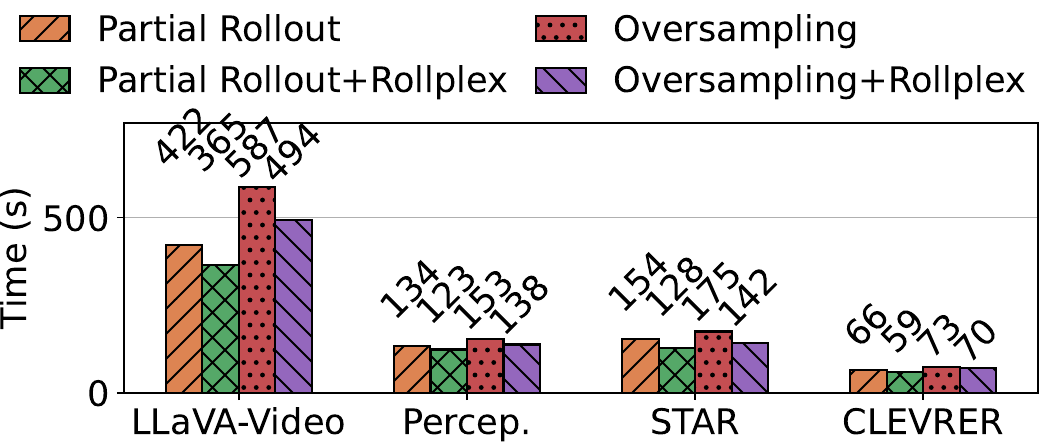}
\caption{Per-step time of partial rollout and oversampling, each with
and without \sysname.}
\label{fig:eval-async}
\end{figure}

On the same 32\,B/32$\times$H800 setup as \S\ref{sec:eval-e2e},
Figure~\ref{fig:eval-async} shows that \sysname's benefit carries over
to partial rollout and oversampling~\cite{april,rollpacker}. Applied
on top, it reduces per-step time by
$1.08\times$--$1.21\times$ for partial rollout and
$1.04\times$--$1.23\times$ for oversampling. Absolute step times are
not comparable across methods because each changes the amount of
rollout work per step, so we compare each method only with and without
\sysname. The gain is slightly smaller than over the plain colocated
baseline because tail-cutting shortens the decode window and leaves
less room to hide prefix work.

\section{Discussion}
\label{sec:discuss}

\PHB{Applicability Envelope.} \sysname applies when substantial
training-side work is a \emph{response-agnostic prefix} that depends on
the prompt and snapshot~$\theta_k$, but not the rollout response. VLM
RL fits this regime because video encoding and visual-token prefill can
materialize prefix KV before the response is known
(\S\ref{sec:workload-iter}). Benefits are largest when prefix work
rivals decode and the GPU budget cannot support separate rollout and
training pools at their preferred TP degrees; they diminish when the
prefix is small or a clean split is affordable.

\PHB{Embodied RL.} VLMs ground agents visually, while
vision-language-action models map observations and instructions to
robot actions~\cite{palme,rt2,openvla}.  Their trajectories create long
visual prefixes from observations, goals, and feedback.  When a fixed
history is response independent, \sysname can harvest decode slack as
in our VLM workloads.  Closed-loop deployment additionally requires
trajectory-aware KV ownership because later observations depend on
earlier actions.

\PHB{Fault Tolerance and Blast Radius.} \sysname relies on soft
GPU sharing through separate CUDA contexts, shared IPC mappings,
streams, and MPS. These mechanisms multiplex kernels and memory but
do not provide independent recovery boundaries. If either colocated
engine triggers a fatal device error, invalidates an IPC mapping, or
wedges a collective, the shared pool will likely lose the iteration.
Disaggregation can instead place the engines in separate pools,
enabling independent restart and failure containment. This limitation
is orthogonal to \sysname's overlap schedule and memory-lifecycle
design; we leave fault isolation and recovery for future work.

\section{Related Work}
\label{sec:related}

\subsection{RL Post-Training Execution}

LLM RL systems orchestrate distributed training and generation
~\cite{deepspeedchat,openrlhf,verl,roll,nemoaligner,slime}.  They use
temporal colocation~\cite{verl,roll,rlhfuse}, disaggregation
~\cite{openrlhf,rollmux,streamrl,tensorhub}, or asynchronous execution
that relaxes rollout--update ordering~\cite{areal,relax}.

\sysname instead spatially overlaps independent work from the
\emph{same} synchronous iteration.  Unlike training systems that
pipeline, co-schedule, or rematerialize work within training
~\cite{dapple,fastermoe,mario}, it crosses rollout and training
engines.  It also lets resident engines read one snapshot at different
TP degrees rather than converting layouts or transferring snapshots
~\cite{verl,tensorhub}.

\subsection{GPU Spatial Sharing}

NVIDIA MPS overlaps kernels and memory copies across processes;
MIG creates isolated compute/memory partitions~\cite{cudamps,nvidiamig}.
Neither chooses
co-runners~\cite{case}.  Research systems add memory
sharing~\cite{salus}, preemption~\cite{reef},
profile-guided co-scheduling~\cite{orion}, compilation-aware
scheduling~\cite{paella}, or dynamic compute-and-bandwidth
regulation~\cite{zipbatch,sgdrc}.

Prior systems multiplex independent jobs for isolation, fairness, or
throughput.  \sysname uses
MPS but derives co-runners from the on-policy dependence graph and
coordinates autograd state, KV caches, optimizer residency, and weights
shared across training and serving.  NanoFlow overlaps operations
inside one serving engine~\cite{nanoflow}; \sysname instead
overlaps models and engines across RL phases.

BEEMS smooths vision-inference memory demand, while vAttention and
GMLake use VMM for stable KV addresses and reduced fragmentation
~\cite{beems,vattention,gmlake}.  \sysname instead uses VMM for
phase-scoped memory residency and cross-process weight views.

\subsection{Rollout Underutilization}

APRIL~\cite{april} and RollPacker~\cite{rollpacker} fill decode slack
with more rollout work.  Serving systems instead co-schedule prefill
with decode~\cite{orca,sarathiserve} or disaggregate the two
~\cite{splitwise,distserve,mooncake,dynamo,semipd,windserve}, using
other requests to improve serving.  \sysname instead fills the window
with response-independent reference and training prefixes from the
same iteration, the dominant exposed cost in VLM RL
(\S\ref{sec:workload-time}), without changing sampled responses or the
PPO/GRPO data dependence.

\subsection{Prefix Sharing}

Prefix sharing~\cite{prefixgrouper,tree_training,areal_dta} removes
redundant prompt-side computation across trajectories sharing a prompt,
as in GRPO, where the $N$ sampled responses in a group share the same
video-text prefix.
\sysname implements the same semantics in all baselines: each role
computes its GRPO prefix once per prompt group and reuses it across the
sampled responses.
Thus, our speedups do not come from eliminating redundant prefix
recomputation in the baselines, but from moving the already-deduplicated,
response-independent prefix work into the rollout-decode window and
preserving its KV state for later scoring and training.


\section{Conclusion}
\label{sec:conc}

In this paper, we present \sysname, a runtime that overlaps VLM
prefixes with rollout decode via phase-aware memory management and
parallelism-aware weight sharing.  On 32$\times$H800 GPUs, it speeds
up serial colocation by $1.23\times$--$1.30\times$ and disaggregation
by $1.57\times$--$2.24\times$ under the same GPU budget.

\PHB{Acknowledgments}
The authors used ChatGPT/Codex for paper refinement and figure generation. All data are collected and verified by human authors. All text and figures are verified by human authors.

\clearpage

\bibliographystyle{ACM-Reference-Format}
\bibliography{references}

\clearpage
\appendix
\section{Appendix}
\label{app:design-details}

This appendix contains detailed implementation contracts that are useful
for reproducing \sysname but are not necessary for following the main
execution schedule.

\subsection{Intermediate-data lifecycle}
\label{app:idata-lifecycle}

Table~\ref{tab:idata-lifecycle} gives the complete per-object state
contract behind the lifecycle policy of \S\mainref{sec:design-mem}: every
tensor class whose lifetime spans a phase boundary of the
\sysname iteration, where it is produced and consumed, whether it is
retained on the autograd graph, and where it resides in the memory
hierarchy. The \emph{Produced}/\emph{Consumed} columns define each
object's lifetime against the three-phase schedule of
\S\mainref{sec:design-overview}; \emph{Autograd} records whether the object
must stay attached to the backward graph for gradient correctness;
\emph{Residency policy} states whether it is pinned in HBM, offloaded to
the host-side pinned pool, or streamed in chunks. The final column marks
the objects that are simultaneously live during the Phase~1 overlap
window and therefore determine whether colocated rollout and prefix
computation fit within HBM.

The rows fall into four groups. First, \emph{inference-side state}
(rollout KV) is HBM-resident only while generation runs and carries no
autograd obligation. Second, \emph{boundary state} (the two boundary
prefix KVs and the prefix logprobs) is the small, latency-critical
output of the Phase~1 prefix forwards that Phase~2 consumes; it stays in
HBM across the phase boundary, and only the training actor's copy is
attached, since gradients re-enter the prefix through that KV boundary.
Third, \emph{bulky training state} (training-prefix activation and
checkpoint state, the gradient buffer, and FP32 master weights with Adam
moments) is exactly what \sysname refuses to keep resident: activation
state is moved out of HBM the moment it is produced, by offloading to the
pinned pool or by discarding for recomputation
(\S\mainref{sec:design-mem-pool}), and rematerialized for the
Phase~2 backward, gradients are offloaded as chunks finalize, and
optimizer state is streamed one chunk at a time during Phase~3. Fourth,
\emph{regenerable or frozen state} (the BF16 actor snapshot and the
reference-model parameters) carries no autograd obligation and can be
discarded and rebuilt---the old snapshot is discarded at the optimizer
barrier and the next one regenerated chunk by chunk from the FP32 master
weights during the Phase~3 stream, and the frozen reference weights are
paged into HBM only around their Phase~1 prefix and Phase~2 scoring
windows.

The final column then makes the overlap-feasibility argument of
\S\mainref{sec:design-mem-contract} concrete: the Phase~1 peak comprises
only rollout KV, the boundary KVs, the shared actor snapshot, the
transiently resident reference weights, and small residuals (prefix
logprobs and offload handles)---while suffix
activations, gradients, and optimizer state contribute nothing to the
Phase~1 budget. This is the per-object justification for why the
combined working set fits where the naive colocation of
\S\mainref{sec:existing-naive} does not.

\begin{table*}
\centering
\caption{Intermediate-data state contract for one \sysname iteration
(P1/P2/P3 = the three phases of \S\mainref{sec:design-overview}).
\emph{Autograd} states whether the object is retained on the
Megatron/PyTorch backward graph; \emph{Residency} gives the phase-scoped
policy. The rightmost column marks the objects that count toward the
Phase~1 overlap-window peak. Phase~2/3-only
state does not inflate the Phase~1 budget. The training-actor prefix
produces attached boundary KV and activation state. The activation state is
offloaded immediately and rematerialized or recomputed for backward, which
preserves the prefix backward path without keeping the full forward tape
resident in HBM.}
\label{tab:idata-lifecycle}
\footnotesize
\setlength{\tabcolsep}{5pt}
\renewcommand{\arraystretch}{1.25}
\begin{tabular}{@{}p{3.0cm}p{1.5cm}p{1.6cm}p{2.0cm}p{4.6cm}c@{}}
\toprule
Object & Produced & Consumed & Autograd & Residency policy & In P1 peak \\
\midrule
Rollout KV & P1 prefill & P1 decode & n/a (inference) & HBM; released after generation & yes \\
Boundary prefix KV (training actor) & P1 prefix fwd & P2 suffix fwd/bwd & \textbf{attached} & HBM; kept until suffix consumes it & yes \\
Boundary prefix KV (reference) & P1 prefix fwd & P2 scoring & detached & HBM; kept until scoring consumes it & yes \\
Reference-model parameters & permanent (frozen) & P1 ref prefix / P2 ref scoring & n/a (frozen) & PinnedPool; loaded to HBM around reference compute, offloaded when it completes & yes (during ref prefix) \\
Prefix logprobs & P1 prefix fwd & P2 scoring/training & detached or attached by role & HBM or host; retained until combined with suffix logprobs & yes (small) \\
Training-prefix activation/checkpoint state & P1 prefix fwd & P2 backward (reload or recompute) & attached & Offloaded immediately to PinnedPool or dropped for recompute; rematerialized during backward & yes (handles) \\
Suffix activations & P2 suffix fwd & P2 backward & attached & HBM (checkpointed); P2 working set & P2-only \\
Gradient buffer & P2 backward & P3 optimizer & leaf grads & Offloaded to host as chunks finalize; re-admitted per chunk in P3 & no \\
BF16 actor snapshot $\theta_k$ & P3 cast (prev iter) & P1/P2 fwd+bwd & n/a (regenerable) & HBM; old snapshot discarded at optimizer barrier, regenerated from FP32 & yes \\
FP32 master weights + Adam moments & permanent & P3 optimizer & n/a & PinnedPool; streamed one chunk at a time in P3 & no \\
\bottomrule
\end{tabular}
\end{table*}

\end{document}